\documentclass[sigconf]{acmart}

\AtBeginDocument{%
  }
\usepackage{multirow}
\setcopyright{acmlicensed}
\copyrightyear{2025}
\acmYear{2025}
\acmDOI{XXXXXXX.XXXXXXX}
\acmConference[NTCIR '25]{}{June 10--13, 2025}{Tokyo, Japan}
\acmISBN{978-1-4503-XXXX-X/2018/06}

\begin{document}

\title{Overview of the NTCIR-18 Automatic Evaluation of LLMs (AEOLLM) Task}




\author{Junjie Chen}
\affiliation{%
  \institution{DCST, Tsinghua University}
  \department{Quan Cheng Laboratory}
  \city{Beijing 100084}
  \country{China}
}
\email{chenjj24@mails.tsinghua.edu.cn}

\author{Haitao Li}
\affiliation{%
  \institution{DCST, Tsinghua University}
  \department{Quan Cheng Laboratory}
  \city{Beijing 100084}
  \country{China}
}
\email{liht22@mails.tsinghua.edu.cn}

\author{Zhumin Chu}
\affiliation{%
  \institution{DCST, Tsinghua University}
  \department{Quan Cheng Laboratory}
  \city{Beijing 100084}
  \country{China}
}
\email{chuzm19@mails.tsinghua.edu.cn}

\author{Yiqun Liu}
\affiliation{%
  \institution{DCST, Tsinghua University}
  \department{Zhongguancun Laboratory}
  \city{Beijing 100084}
  \country{China}
}
\email{yiqunliu@tsinghua.edu.cn}

\author{Qingyao Ai}
\affiliation{%
  \institution{DCST, Tsinghua University}
  \department{Quan Cheng Laboratory}
  \city{Beijing 100084}
  \country{China}
}
\email{aiqy@tsinghua.edu.cn}


\begin{abstract}
In this paper, we provide an overview of the NTCIR-18 Automatic Evaluation of LLMs (AEOLLM) task. As large language models (LLMs) grow popular in both academia and industry, how to effectively evaluate the capacity of LLMs becomes an increasingly critical but still challenging issue. Existing methods can be divided into two types: manual evaluation, which is expensive, and automatic evaluation, which faces many limitations including task format (the majority belong to multiple-choice questions) and evaluation criteria (occupied by reference-based metrics). To advance the innovation of automatic evaluation, we propose the AEOLLM task which focuses on generative tasks and encourages reference-free methods. Besides,  we set up diverse subtasks such as dialogue generation, text expansion, summary generation and non-factoid question answering to comprehensively test different methods. This year, we received 48 runs from 4 teams in total. This paper will describe the background of the task, the data set, the evaluation measures and the evaluation results, respectively.
\end{abstract}




\maketitle

\section{Introduction}
\label{section one}
\label{section: intro}
The Automatic Evaluation of LLMs (AEOLLM) task is a core task in NTCIR-18 to support in-depth research on large language models (LLMs) evaluation. 

In recent times, the persistent advancement of LLMs like DeepSeek-V3 \cite{liu2024deepseek}, GPT-4\cite{achiam2023gpt}, Qwen~\cite{bai2023qwen, yang2024qwen2} and Chatglm \cite{glm2024chatglm} has sparked a lot of interest in Artificial General Intelligence within both academia and industry. However, the rapid advancement of LLMs has introduced a key challenge in the progression of these models\textemdash\textemdash efficiently and effectively evaluating their performance. A dependable and reusable LLMs evaluation method can not only aid in the judicious selection of the best LLMs for specific tasks but also impart valuable insights for optimizing LLMs.

The existing LLM evaluation methods could be categorized into two groups: manual evaluation and automatic evaluation. Manual evaluation involves engaging human annotators to directly assess the quality of responses generated by LLMs. Conversely, automatic evaluation relies on standard metrics and evaluation tools to assess model performance. In comparison to human evaluation, automatic evaluation mitigates the need for extensive human involvement, thereby reducing costs and enhancing the objectivity of the evaluation process \cite{chang2023survey}. As a result, investigating efficient and general automatic evaluation methods becomes increasingly significant and holds great promise.

However, existing automatic evaluation methods for LLMs still have the following limitations: (1) Limited task format. The majority of existing automatic LLM evaluation benchmarks only contain multiple-choice-format questions. Although this setting facilitates the computation of evaluation metrics, the multiple-choice format differs from the real-world practical questions, which are usually open-ended without definite answers; (2) Limited evaluation criteria. Reference-based metrics~\cite{lin2004rouge, papineni2002bleu, zhang2019bertscore} (such as Rouge and BLEU) are widely used in many LLM tasks. These kinds of metrics cannot accurately reflect the quality of the results. Moreover, LLMs' strong memory capabilities enable them to rapidly learn and optimize any public benchmarks during training, which makes the reference-based metrics on these benchmarks useless for testing immediately.

Based on these considerations, we propose the NTCIR-18 Automatic Evaluation of LLMs (AEOLLM) task, which: (1) concentrates on generative tasks to evaluate the capacity of automatic evaluation methods for LLMs in assessing open-ended responses, and (2) encourages participants to develop reference-free evaluation methods. To make our task more comprehensive, we set up multiple types of tasks including dialogue generation, text expansion, summary generation and non-factoid question answering. We believe that this task will assist in improving the performance of LLM evaluation methods and significantly advance the development of LLMs.

\begin{table}[]
\caption{NTCIR-18 AEOLLM timeline (time zone: AOE).}
\label{timeline}
\begin{tabular}{ll}
\hline
Dataset Release            & May 1, 2024              \\
Dry run Deadline           & Jan 15, 2025             \\
Formal run        & Jan 15, 2025-Feb 1, 2025 \\
Evaluation Results Release & Feb 1, 2025              \\ \hline
\end{tabular}
\end{table}

\begin{table}[]
\caption{NTCIR-18 AEOLLM run statistics.}
\label{run}
\begin{tabular}{lccc}
\hline
Team             & Dry run & Formal run & Total \\ \hline
KNUIR            & 7       & 2          & 9     \\
ISLab            & 18      & 3          & 21    \\
UCLWI            & 1       & 1          & 2     \\
PanguIR & 10      & 6          & 16    \\ \hline
Total            & 36      & 12         & 48    \\ \hline
\end{tabular}
\end{table}

Timeline of the NTCIR-18 AEOLLM task is shown in Table \ref{timeline}. This year we received 48 runs from 4 teams in total. The statistics are given in Table \ref{run}. 

In summary, our contributions include: (1) Providing a practical dataset. (2) Offering an automated evaluation pipeline that facilitates performance testing of different methods. More details can be found at \url{https://huggingface.co/spaces/THUIR/AEOLLM}.

The structure of the remaining sections of this paper is described as follows. In Section 2, we present the overall design and evaluation framework for AEOLLM. In Section 3, we detail the implementation, including dataset construction and evaluation metrics. In Section 4, we present the final evaluation results. Finally, we conclude in Section 5.

\section{Task Framework}
\begin{figure}[t]
  \centering
  \includegraphics[width=\columnwidth]{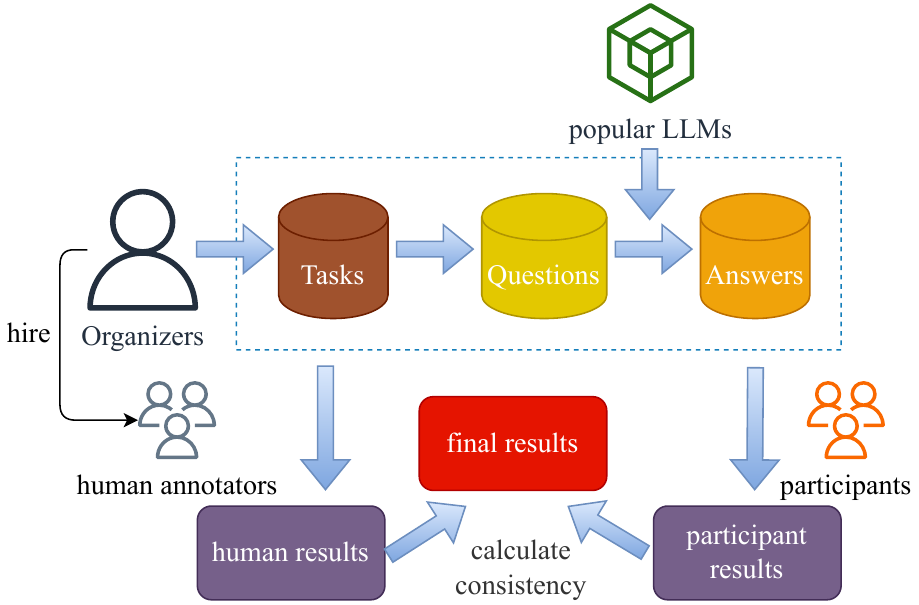}
  \caption{The overall framework of the AEOLLM task.}
  \label{all_frame}
\end{figure}

Figure \ref{all_frame} illustrates the overall framework of the AEOLLM task. As discussed in Section 1, we focus on open-ended tasks. Therefore, we first select four representative generative tasks: 

\textbf{Dialogue generation} requires generating human-like responses to numerous topics in daily conversation contexts. For this, we select DailyDialog \cite{li2017dailydialog}, which is a high-quality dataset of 13k multi-turn dialogues with less noise.

\textbf{Text Expansion} requires participants to generate stories based on the given theme. We select Writingprompts \cite{fan2018hierarchical} as the dataset, which is a large dataset of 300K human-written stories paired with writing prompts from an online forum.

\textbf{Summary generation} involves generating a summary for a given text. For this task, we utilize the Extreme Summary (Xsum) dataset \cite{narayan-etal-2018-dont}, which consists of over 220,000 real single-document news and summaries collected by the British Broadcasting Corporation (BBC).

\textbf{Non-factual question-answering} refers to providing answers to questions that do not have fixed responses. For this task, we choose the NF\_CATS dataset \cite{bolotova2022non}. It contains about 12,000 non-factual questions and these questions are categorized into eight classes to differentiate the difficulty levels.

Second, we choose a series of popular LLMs during the competition to generate answers. Then we manually annotate the answer sets for each question, which will be used as gold standards for evaluating the performance of different evaluation methods.

Last, we collect evaluation results from participants on Answer sets for different Question sets under each task, and calculate consistency with manually annotated results. We use accuracy ($acc$), Kendall's tau ($\tau$) \cite{kendall1938new} and Spearman correlation coefficient ($\rho$) \cite{lehman2013jmp}  as the evaluation metrics, which will be discussed next.

\section{Implementation Details}
\subsection{Datasets}

\subsubsection{Answer Generation} We randomly sampled 100 instances from each dataset as the question set and selected 7 different LLMs to generate answers, forming the answer set. 

As a result, each dataset produced 700 instances, totaling 2,800 instances across the four datasets.

\subsubsection{Human Annotation} For each instance (question-answer pair), we employed human annotators to provide a score ranging from 1 to 5 and took the median of these scores as the final score.

Based on this score, we calculated the rankings of the 7 answers for each question. If scores were identical, the answers were assigned the same rank, with the lowest rank being used.

\subsubsection{Data Acquisition and Usage} We divided the 2,800 instances into three parts: 

1. train set: 20\% of the data (covering all four datasets) was designated as the training set (including human annotations) for participants to reference when designing their methods.

2. test set: Another 20\% of the data was set aside as the test set (excluding human annotations), used to evaluate the performance of participants' methods and to generate the leaderboard.

3. reserved set: The remaining 60\% of the data was reserved for the final evaluation.

Both the training set and the test set can be downloaded from the provided link: \url{https://huggingface.co/datasets/THUIR/AEOLLM}.

\subsection{Evaluation Metrics}
\subsubsection{Accuracy ($acc$)} The proportion of identical preference results between the model and human annotations. Specifically, we first convert individual scores (ranks) into pairwise preferences and then calculate consistency with human annotations.

\subsubsection{Kendall's tau ($\tau$)} Measures the ordinal association between two ranked variables.

\[
\tau = \frac{C - D}{\frac{1}{2} n(n - 1)}
\]

where:
\begin{itemize}
    \item \(C\) is the number of concordant pairs
    \item \(D\) is the number of discordant pairs
    \item \(n\) is the number of pairs
\end{itemize}

\subsubsection{Spearman's Rank Correlation Coefficient ($\rho$)} Measures the strength and direction of the association between two ranked variables.

\[
\rho = 1 - \frac{6 \sum d_i^2}{n(n^2 - 1)}
\]

where:
\begin{itemize}
    \item \(d_i\) is the difference between the ranks of corresponding elements in the two lists
    \item \(n\) is the number of elements
\end{itemize}

\subsection{Answer Organization}

We follow a similar format as the ones used by most TREC~\cite{voorhees2005trec} submissions:

\begin{enumerate}
    \item White space is used to separate columns.
    \item The width of the columns in the format is not important, but it is important to have exactly five columns per line with at least one space between the columns.
\end{enumerate}

These five columns are \textbf{taskId questionId answerId score rank}.

\begin{itemize}
    \item The first column is the \textbf{taskId} (index different tasks).
    \item The second column is \textbf{questionId} (index different questions in the same task).
    \item The third column is \textbf{answerId} (index the answer provided by different LLMs to the same question).
    \item The fourth column is \textbf{score} (index the score to the answer given by participants).
    \item The fifth column is \textbf{rank} (index the rank of the answer within all answers to the same question).
\end{itemize}
\section{Participants' Methods}
\begin{itemize}
    \item \textbf{KNUIR} aims to propose automated evaluation methods for LLMs that approximate human judgment by exploring and comparing two distinct approaches: (1) LLM-based scoring, which utilizes GPT models with prompt engineering, and (2) feature-based machine learning, employing transformer-based metrics such as BERTScore, semantic similarity, and keyword coverage. Their findings indicate that LLM-based methods demonstrated scalability but lacked explainability, whereas feature-based approaches provided better interpretability but required extensive tuning, highlighting the trade-offs between the two strategies. 
    \item \textbf{ISLab}  leverages data augmentation to increase the volume of training data and employs the odds ratio preference optimization (ORPO) algorithm for reinforcement learning to optimize the evaluator.
    \item \textbf{UCLWI} proposes an efficient evaluation pipeline for Retrieval-Augmented Generation (RAG) systems tailored for low-resource settings. Their method uses ensemble similarity measures combined with a logistic regression classifier to assess answer quality from multiple system outputs using only the available queries and replies.
    \item \textbf{PanguIR}~\cite{mei2025panguir} proposes three key
methods to improve the reference-free evaluation: (1) Multi-model Collaboration: Leveraging multiple LLMs to approximate human ratings across various subtasks; (2) Prompt Auto-optimization: Utilizing LLMs to iteratively refine the initial task prompts based on evaluation feedback from training samples; and (3) In-context
Learning (ICL) Optimization: Based on the multi-task evaluation feedback, the authors train a specialized in-context example retrieval model, combined with a semantic relevance retrieval model, to jointly identify the most effective in-context learning examples.

\end{itemize}
\section{Official Results}
\begin{table*}[t]
\caption{The results from the formal run on the reserved set. Baselines 1, 2, 3, and 4 correspond to direct prompting of ChatGLM3-6B, Baichuan2-13B, ChatGLM-Pro, and GPT-4o, respectively. The best result is highlighted in bold.}
\label{result}
\resizebox{\textwidth}{!}{%
\begin{tabular}{cccccccccccccccc}
\hline
\multirow{2}{*}{\textbf{Team}}                  & \multicolumn{3}{c}{\textbf{Dialogue Generation}}    & \multicolumn{3}{c}{\textbf{Text Expansion}}         & \multicolumn{3}{c}{\textbf{Summary Generation}}     & \multicolumn{3}{c}{\textbf{Non-Factoid QA}}         & \multicolumn{3}{c}{\textbf{Overall}}                \\
                                       & \textbf{$acc$}  & \textbf{$\tau$} & \textbf{$\rho$} & \textbf{$acc$}  & \textbf{$\tau$} & \textbf{$\rho$} & \textbf{$acc$}  & \textbf{$\tau$} & \textbf{$\rho$} & \textbf{$acc$}  & \textbf{$\tau$} & \textbf{$\rho$} & \textbf{$acc$}  & \textbf{$\tau$} & \textbf{$\rho$} \\ \hline
\textbf{Baseline1}                              & 0.5583          & 0.3228          & 0.3495          & 0.5029          & 0.1236          & 0.1293          & 0.5976          & 0.2589          & 0.2712          & 0.6445          & 0.3717          & 0.3948          & 0.5759          & 0.2692          & 0.2862          \\
\multicolumn{1}{l}{\textbf{Baseline2}} & 0.5518          & 0.1647          & 0.1750          & 0.5021          & 0.0698          & 0.0725          & 0.6112          & 0.2693          & 0.2813          & 0.6044          & 0.2664          & 0.2791          & 0.5674          & 0.1926          & 0.2020          \\
\multicolumn{1}{l}{\textbf{Baseline3}} & 0.5924          & 0.2921          & 0.3135          & 0.5495          & 0.2831          & 0.2985          & 0.7079          & 0.4390          & 0.4575          & 0.6975          & 0.4682          & 0.4990          & 0.6368          & 0.3706          & 0.3921          \\
\multicolumn{1}{l}{\textbf{Baseline4}} & 0.6595          & 0.4423          & 0.4797          & 0.5543          & \textbf{0.3963} & \textbf{0.4248} & 0.7029          & 0.3886          & 0.4180          & 0.7441          & \textbf{0.4896} & \textbf{0.5281} & 0.6652          & 0.4292          & 0.4627          \\
\textbf{KNUIR}                         & 0.6778          & 0.4404          & 0.4717          & 0.5512          & 0.3141          & 0.3430          & 0.7375          & 0.4524          & 0.4914          & 0.6951          & 0.4102          & 0.4297          & 0.6654          & 0.4043          & 0.4340          \\
\textbf{ISLab}                         & /               & /               & /               & 0.5241          & 0.3609          & 0.4035          & \textbf{0.7658} & 0.5117          & 0.5632          & /               & /               & /               & /               & /               & /               \\
\textbf{UCLWI}                         & \textbf{0.7756} & \textbf{0.5798} & \textbf{0.6426} & 0.5266          & 0.3482          & 0.3815          & 0.7273          & \textbf{0.5432} & \textbf{0.5763} & 0.6853          & 0.4105          & 0.4291          & 0.6787          & \textbf{0.4704} & \textbf{0.5074} \\
\textbf{PanguIR}                       & 0.7444          & 0.5611          & 0.6091          & \textbf{0.5581} & 0.3432          & 0.3775          & 0.7479          & 0.5097          & 0.5520          & \textbf{0.7528} & 0.4175          & 0.4534          & \textbf{0.7008} & 0.4579          & 0.4980          \\ \hline
\end{tabular}%
}
\end{table*}
Table \ref{result} shows the performance comparison between different baselines and methods from the formal run on the reserved set. Based on the experimental results, the following observations can be made.

\begin{itemize}
    \item \textbf{Comparing different methods,} overall, PanguIR achieves the best performance in terms of accuracy ($acc$), while UCLWI excels in Kendall’s Tau ($\tau$) and Spearman’s Rank correlation coefficients ($\rho$). For each subtask, UCLWI excels in all three metrics for Dialogue Generation and in $\tau$ and $\rho$ for Story Generation. PanguIR outperforms others in $acc$ for Text Expansion and Non-Factual QA, and ISLab leads in $acc$ for Summary Generation. We observe that the best values for different subtasks appear in different methods, suggesting that the selected subtasks are not entirely homogeneous, and indicating that our dataset enables a more comprehensive evaluation. Interestingly, our simple baseline, which directly prompts GPT-4o, achieves the best $\tau$ and $\rho$ in Text Expansion and Non-Factual QA. This highlights that current strong LLMs already demonstrate impressive evaluation performance and show great potential for implementation in LLMs-as-judges~\cite{li2024llms, chen2024automatic, li2024calibraeval}.

    \item \textbf{Comparing different evaluation metrics,} the results of $\tau$ and $\rho$ are almost consistent. This may be because both are measures of concordance, focusing on the relative order between pairs. However, $\tau$ and $\rho$ can sometimes yield an undefined or indeterminate value (NaN) when there are ties in the data, particularly when the two lists have identical values for certain pairs of elements. In contrast, $acc$ sometimes differs from the results of these two coefficients. This suggests that considering multiple metrics is necessary to provide a more comprehensive assessment of the performance of a method.

    \item \textbf{Comparing different subtasks,} the Text Expansion dataset is the most challenging, with the highest $acc$ being only 0.5581. This may be due to the longer length of the answers in story generation (average question length: 27.47, answer length: 311.36, total length: 338.83), making it more difficult for the methods to provide accurate evaluations. This presents a challenging scenario for future method optimization. In contrast, Dialogue Generation is the easiest of the four tasks, likely because these dialogues tend to be shorter (average question length: 99.15, answer length: 16.14, total length: 115.29). These observations indicate that our dataset covers a wide range of difficulties, addressing a variety of application scenarios.
\end{itemize}
\section{Conclusion}
This paper provided an overview of the NTCIR-18 Automatic Evaluation of LLMs (AEOLLM) task, which aims to explore effective methods for automatic evaluation of large language models (LLMs). AEOLLM this year received a total of 48 runs from 4 different teams, showcasing a variety of approaches to evaluating LLMs across four distinct subtasks: dialogue generation, text expansion, summary generation, and non-factoid question answering. Through these evaluations, we observed notable findings: (1) Comparing different methods, overall, PanguIR achieved the best performance in terms of accuracy (acc), while UCLWI excelled in Kendall’s Tau ($\tau$) and Spearman’s Rank correlation coefficients ($\rho$). (2) Considering multiple metrics is necessary to provide a more comprehensive assessment of the performance of a method. (3) The Text Expansion dataset is the most challenging, with the highest $acc$ being only 0.5581. This presents a challenging scenario for future method optimization. Looking ahead, we plan to further extend the AEOLLM task to better and more comprehensively evaluate LLMs.


\bibliographystyle{ACM-Reference-Format}
\bibliography{sample-base}

\end{document}